\documentclass{article}

\usepackage{microtype}
\usepackage{graphicx}
\usepackage{subfigure}
\usepackage{booktabs} % for professional tables
\usepackage{breqn}
\usepackage{comment}

\usepackage[colorlinks=true, linkcolor=black, citecolor=blue, urlcolor=purple]{hyperref}

\PassOptionsToPackage{numbers, compress}{natbib}
\usepackage[preprint]{neurips_2025}

\usepackage{amssymb}
\usepackage{amsmath}
\usepackage{amsthm}
\usepackage{multirow}

\usepackage[dvipsnames]{xcolor}

\definecolor{winter}{rgb}{0.85,0.08,0.2}
\definecolor{summer}{rgb}{0.95,0.53,0.18}
\definecolor{spring}{rgb}{0.02,0.93,0.68}
\definecolor{autumn}{rgb}{0.02,0.68,0.9}

\usepackage{algorithm}
\usepackage{pseudo}

\usepackage{authblk}

\title{From epilepsy seizures classification to detection: \\ A deep learning-based approach for raw EEG signals}

\author[1,2]{Davy Darankoum}
\author[1]{Manon Villalba}
\author[1]{Clélia Allioux}
\author[1]{Baptiste Caraballo}
\author[1]{Carine Dumont}
\author[1]{Eloïse Gronlier}
\author[1]{Corinne Roucard}
\author[1]{Yann Roche}
\author[1]{Chloé Habermacher}
\author[2,*]{Sergei Grudinin}
\author[1,*]{Julien Volle}

\affil[1]{SynapCell SAS, 38330 Saint-Ismier, France}
\affil[2]{Univ. Grenoble Alpes, CNRS, Grenoble INP, LJK, 38000 Grenoble, France}
\affil[*]{These authors contributed equally to this work and share last authorship}

\begin{document}

\maketitle

%%%%%%%%%%%%%%%%%%%%%%%%%%%%%%%%%%%%%%%%%%%%%%%%%%%%%%%%%%%%%%%%%%%%%%%%

\begin{abstract}
Epilepsy is the most prevalent neurological disorder in the world. Although epilepsy has been recognized for centuries, clinical doctors still lack reliable automated tools to diagnose epileptic seizures in electroencephalograms (EEGs). The research community has made significant efforts to develop automated systems for identifying and quantifying epileptic seizures, with many studies reporting excellent accuracy. However, clinicians continue to rely on manual annotations due to the poor generalization performance of automated techniques when applied to EEG data from new patients. Another challenge in the field is translating the results of animal preclinical studies to the clinical ones on humans. 

This work contributes to both challenges. Firstly, we investigate the reasons behind the lack of generalization of automatic models. We find out that while most existing techniques are assessed on seizure classification tasks, clinical doctors face detection tasks in their practice. We demonstrate that the performance of automated pipelines differs significantly between the two and identify the key distinction between the tasks: classification presumes a prior separation between seizure and non-seizure EEG signals, whereas detection requires no such prior knowledge. 
Secondly, we bridge the gap between preclinical and clinical studies by developing novel deep learning architectures. Our best model, trained on EEG data from epileptic mice, demonstrates excellent generalization with an F1-score of 93\% when tested on human data.
\end{abstract}

\begin{center}
\textbf{Corresponding authors:} 
\texttt{sergei.grudinin@univ-grenoble-alpes.fr};
\texttt{jvolle@synapcell.fr}
\end{center}

\vspace{1em}

%%%%%%%%%%%%%%%%%%%%%%%%%%%%%%%%%%%%%%%%%%%%%%%%%%%%%%%%%%%%%%%%%%%%%%%%

\section{Introduction}

%%%%%%%%%%%%%%%%%%%%%%%%%%%%%%%%%%%%%%%%%%%%%%%%%%%%%%%%%%%%%%%%%%%%%%%%

Epilepsy affects more than 50 million individuals worldwide and is characterized by recurring seizures arising from abnormal brain activity, profoundly impacting daily functioning and quality of life \citep{thijs2019epilepsy, beghi2020epidemiology}. Epilepsy can appear through several syndromes, underscoring the need for precise and effective diagnosis to orient epileptic patients toward appropriate healthcare treatments \citep{nabbout2020impact}. Anti-seizure medications (ASMs) suffer from high patient-dependent responses. The development of accurate tools to extract specific information from epileptic patients appears then to be paramount for developing better-suited medications.

Electroencephalogram (EEG) stands as a pivotal tool in epilepsy diagnosis, thanks to one of its capacities to capture substantial alterations in brain electrical activity during and in proximity to epileptic seizures \citep{noachtar2009role,mesraoua2019eeg,shoji2021automated}. Neurologists classify brain activity into four distinguishable phases based on EEG inspection. The preictal phase represents the period preceding a seizure; the ictal phase corresponds to the actual seizure event; the postictal phase encompasses the time following a seizure episode; and the interictal phase constitutes the interval between seizure occurrences, distinct from the other states. Different analysis processes involving some or all of these four phases help neurologists diagnose the proper type of epilepsy \citep{toraman2021automatic}. Moreover, epileptic seizure identification and quantification are broadly used to evaluate the efficacy of new ASMs and disease-modifying therapies.

Among the different epilepsy syndromes, Mesial Temporal Lobe Epilepsy (MTLE), characterized by refractory seizures, is the most common type of focal epilepsy in adults. Approximately 30-50\% of patients with MTLE develop drug resistance \citep{paschen2020hippocampal,ammothumkandy2022altered}. Considerable research endeavors presently focus on optimizing the preclinical stage of drug development for epilepsy to enhance translational success and improve the likelihood of therapeutic candidates advancing through clinical trials. The preclinical phase designates a stage where studies are conducted on laboratory animals to identify the best treatment candidates among many and to determine safe doses in order to accelerate and increase success chances on the tests carried out on humans, namely the "clinical phase". The translational properties of EEG signals make them a valuable tool for monitoring brain activity in animal models, facilitating the extrapolation of findings to human brain function. However, due to a lack of accurate analysis tools, neurologists often review and interpret EEG signals manually, which leads to misidentification of epileptic seizures, inefficiencies, and subjectivity. Therefore, there is a need to develop automated techniques to identify seizures accurately and minimize diagnostic errors. This automation task is rather challenging given the complex characteristics of EEG signals, including their low signal-to-noise ratios, high-frequency dimension, non-stationarity, non-linearity, variability, and the presence of artifacts. 

Thanks to the constant progress in machine learning-based techniques, many repetitive data annotation tasks can now be automated, thus affording minimal room for errors and liberating oneself from protracted, time-consuming activities \citep{ebrahim2020quantitative,jumper2021alphafold,lecun2015deep,min2016deep}. These advances have also extended to seizure detection in EEG signals \citep{chen2023automated,choi2019novel,hussain2019epileptic}. We can divide approaches that automate epileptic seizure detection through machine learning into two main categories. The first includes research focusing on hand-crafted feature extraction from EEG signals, followed by training classical machine learning or deep learning models to result in an epileptic seizure detection tool \citep{mahjoub2020epileptic, shoeb2010application}. The second category tackles the task by training deep learning models to automatically extract meaningful features in EEG signals and simultaneously perform seizure detection. 
One can manually extract features from the EEG signals in multiple ways: in the time domain, frequency domain, time-frequency domain, or with nonlinear analysis \citep{durongbhan2019dementia,singh2023trends}. A study introduced by \citet{guo2010automatic} presents an epileptic seizure detection pipeline based on the computation of line-length features from wavelet transform-based signal decomposition, a technique that allows feature extraction from the time-frequency domain of EEG signals. Then, the authors used the extracted features with a Multi-Layer Perceptron Neural Network (MLPNN) to perform seizure detection. Their model achieved very high performance (98\% accuracy on the classification into seizure and non-seizure segments) on the Bonn University dataset published by \citet{andrzejak2001indications}. In another study, \citet{wang2018hardware} proposed a real-time seizure detection algorithm based on Short Time Fourier Transform (STFT), another technique to extract time-frequency domain features from the EEG, and then trained them with a Support Vector Machine (SVM), a machine learning-based model. The authors tested their pipeline on the CHB-MIT Scalp EEG database and achieved 98\% sensitivity on the seizure detection task. \citet{mursalin2017automated} performed a correlation-based feature selection from the time domain and frequency domain of EEGs and then applied an ensemble of random forest classifiers (machine learning-based models) to detect seizures on EEG. On the Bonn University dataset, their model achieved 97\% accuracy. Despite the high reported accuracies, the models developed for epileptic seizure detection and based on manual feature extraction exhibit several limitations, mainly regarding their capacity to generalize across diverse subject profiles and conditions \citep{ebrahim2020quantitative}. This poor generalization indicates that some features invariant to variabilities across subjects’ EEG signals might not be learned in the trained models. Moreover, the subject-dependent signal-to-noise ratio causes a high variability of feature importance for seizure detection. Some researchers attempted to overcome these limitations by developing deep learning models that automatically extract meaningful features in EEG signals by learning invariant embeddings. 

Deep learning models, such as Convolutional Neural Networks (CNNs), with their capacity for extracting local features, and Recurrent Neural Networks (RNNs), capable of capturing long-range relationships, serve as relevant tools for acquiring consistent and translational features essential for EEG classification into seizure and non-seizure segments. \citet{acharya2018deep} proposed one of the first CNN-based networks trained on raw time series EEG to classify EEG into seizure and non-seizure activities. They achieved 89\% accuracy on the Bonn University dataset. \citet{roy2018deep} applied different EEG pre-processing techniques coupled to several neural network architectures, namely, a 1D CNN, a 2D CNN, and a 1D CNN-GRU (Gated Recurrent Unit), to classify EEG signals into normal and abnormal activities. Their best model (1D-CNN-GRU) demonstrated 99\% accuracy on the TUH Abnormal EEG Corpus \citep{obeid2016tuh}. \citet{cho2020comparison}  compared four input modalities (raw time series EEG, periodograms that reflect the spectral density of EEG signals, 2D images from STFT coefficients, and 2D images from raw EEG waveforms) and different neural networks for an epileptic seizure detection task. They trained fully connected neural networks, RNNs and CNNs, on unique or combined input types listed above. Their best pipeline led to 99\% accuracy on UPenn and Mayo Clinic’s Seizure Detection challenge datasets.

This study presents the development of deep learning models to automatically detect epileptic seizures in EEG signals from an animal model of MTLE, the intra-hippocampal kainate mouse model. We focused on MTLE mainly due to its prevalence worldwide \citep{tatum2012mesial} and because MTLE manifests by seizures occurring in a specific brain region, therefore limiting the use of common multi-electrode set-ups for EEG recording. MTLE is a focal epilepsy type characterized by recurrent seizures with an onset involving the amygdalohippocampal complex and parahippocampal region. Consequently, seizures can be captured only using a single electrode positioned near the onset site. In contrast, non-focal epilepsy types involve seizures occurring across multiple brain regions, allowing for the use of multiple electrodes to capture them. Our second goal was the development of accurate tools for enhancing the preclinical studies workflow. We trained and validated the developed neural network architectures using EEG signals recorded on MTLE mice. We evaluated the generalization performance of our top-performing models by applying them to signals from human patients. Our exploration of different neural network architectures included convolutional neural networks, recurrent neural networks, segmentation models based on the U-Net architecture, and Attention-based networks. We further developed a post-processing algorithm that concatenates overlapping signal segments into a continuous time series, which allowed the detection of seizures in a real-world scenario. Finally, we identified two evaluation strategies for assessing model performance, which we believe are more effective for benchmarking the efficacy of automated seizure detection tools.

%%%%%%%%%%%%%%%%%%%%%%%%%%%%%%%%%%%%%%%%%%%%%%%%%%%%%%%%%%%%%%%%%%%%%%%%
\section{Materials and Methods}
\subsection{Animals - MTLE mice model}
Animal experiments were approved by the ethical committee of the Grenoble Institute of Neuroscience, University Grenoble Alpes, and performed by SynapCell in accordance with the European Committee Council directive of September 22, 2010 (2010/63/EU). 
\citet{duveau2017mesiotemporal} has previously described detailed information about the generation of the MTLE mouse model. Briefly, adult male C57Bl/6J mice (11 weeks of age) receive a kainic acid injection in the right dorsal hippocampus (AP = -2, ML = -1.5, DV = -2 mm relative to bregma) \citep{paxinos2019paxinos}. During the surgical procedure, a bipolar electrode is positioned in the right dorsal hippocampus (AP = -2.4, ML = -1.5, DV = -2 mm relative to bregma). The implant is secured to the skull using dental cement to allow tethered EEG recordings in freely moving animals. After surgery, animals are left in their home cage for at least one week of recovery. After the epileptogenesis period lasting four weeks, mice became accustomed to the recording conditions, and EEGs were recorded to assess each animal. Different criteria (number of HPD = Hippocampal Paroxysmal Discharges, sufficient signal-to-noise ratio), allowing a distinct determination of the beginning and end of events, are used to enroll animals in a study. For all experiments, the criterion of inclusion is rigorously the same. During the study, animals are connected to an amplifier by a recording cable that does not restrict their movement. The EEG signal is band-pass filtered between 0.8 Hz and 1 kHz and digitized at 512 Hz (SDLTM128 Channels; Micromed, France). EEGs are stored for offline analysis, allowing experts to evaluate all the EEG traces and annotate the boundaries of each HPD.

\subsection{Datasets}
In our study, we used two datasets. Both contain EEG signals recorded on subjects (mice for the first dataset and humans for the second one) who suffer from MTLE.

\begin{figure}
    \centering

    \includegraphics[width=1\linewidth]{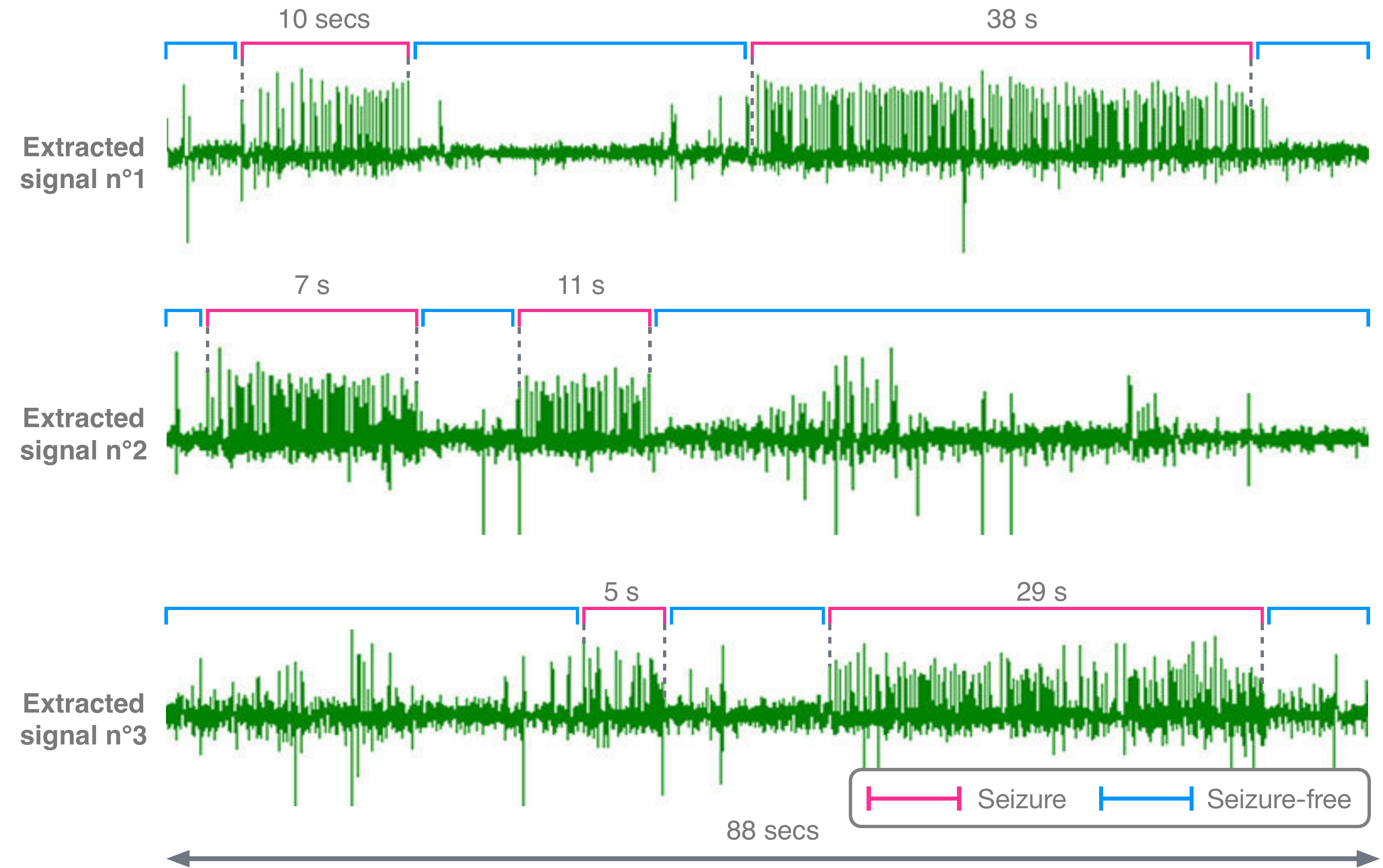}
    \caption{Examples of EEG signals from Dataset 1. Snapshots of 3 portions of EEG signals
    measured in 3 mice and labeled by the same expert. Labels in red indicate detected seizures,
    and blue labels represent seizure-free activities. The total duration of the snapshots is 88
    seconds.
    }
    \label{fig:EEG_samples}
\end{figure}

\paragraph{Dataset 1:} This dataset results from a selection of EEGs recorded in ten different studies conducted at SynapCell (Figure \ref{fig:EEG_samples}). It includes 1440 hours of EEG signals recorded in 136 MTLE mice: 1190 hours of seizure-free activity and 250 hours of epileptic seizures. Each mouse is recorded following an average of 3 different sessions, with each session lasting approximately 3 hours. Each signal has been reviewed by an expert scorer, who labeled ictal activities with the assistance of commercial software for seizure detection (Deltamed Coherence, Natus Medical Incorporated, USA). 

\paragraph{Dataset 2:} The second dataset is public from Bonn University \citep{andrzejak2001indications}. 
The dataset contains EEG signals recorded on healthy humans and humans suffering from MTLE through a multi-channel set-up, but the authors provide only one-channel data. We chose it for our study as it exhibits the same EEG data constraints of single-channel measurements as Dataset 1. It also allows assessing the generalization abilities of our models from the preclinical setting to the clinical environment. 
The dataset comprises five different sets, denoted from A to E. Each contains 100 EEG segments of 23.6 sec. These segments result from continuous EEG recordings that were processed by the authors to remove artifacts. In total, the authors recorded EEG signals from ten humans: five healthy and five diagnosed with MTLE. The EEG signals recorded through the head surface of the five healthy volunteers formed sets A and B. Set A consists of EEGs recorded with eyes open and set B with eyes closed. Presurgical EEGs from five patients suffering from MTLE were used to constitute sets C, D, and E. Set D comprises EEGs recorded from the epileptogenic zone. Set C comprises EEGs measured from the hippocampal structure of the opposite hemisphere. Segments in sets C and D contain only the brain activity measured during seizure-free intervals, whereas set E contains EEG segments recorded through the epileptogenic zone (the hippocampal structure on the hemisphere from which the seizures originate) during seizure activity. The authors recorded all the EEG signals from these sets at a sampling rate of 173.61 Hz and a bandpass filter to keep only frequencies between 0.53 and 40 Hz.

\subsection{Pre-processing and post-processing pipelines}

\begin{figure}
    \centering
    \includegraphics[width=0.6\linewidth]{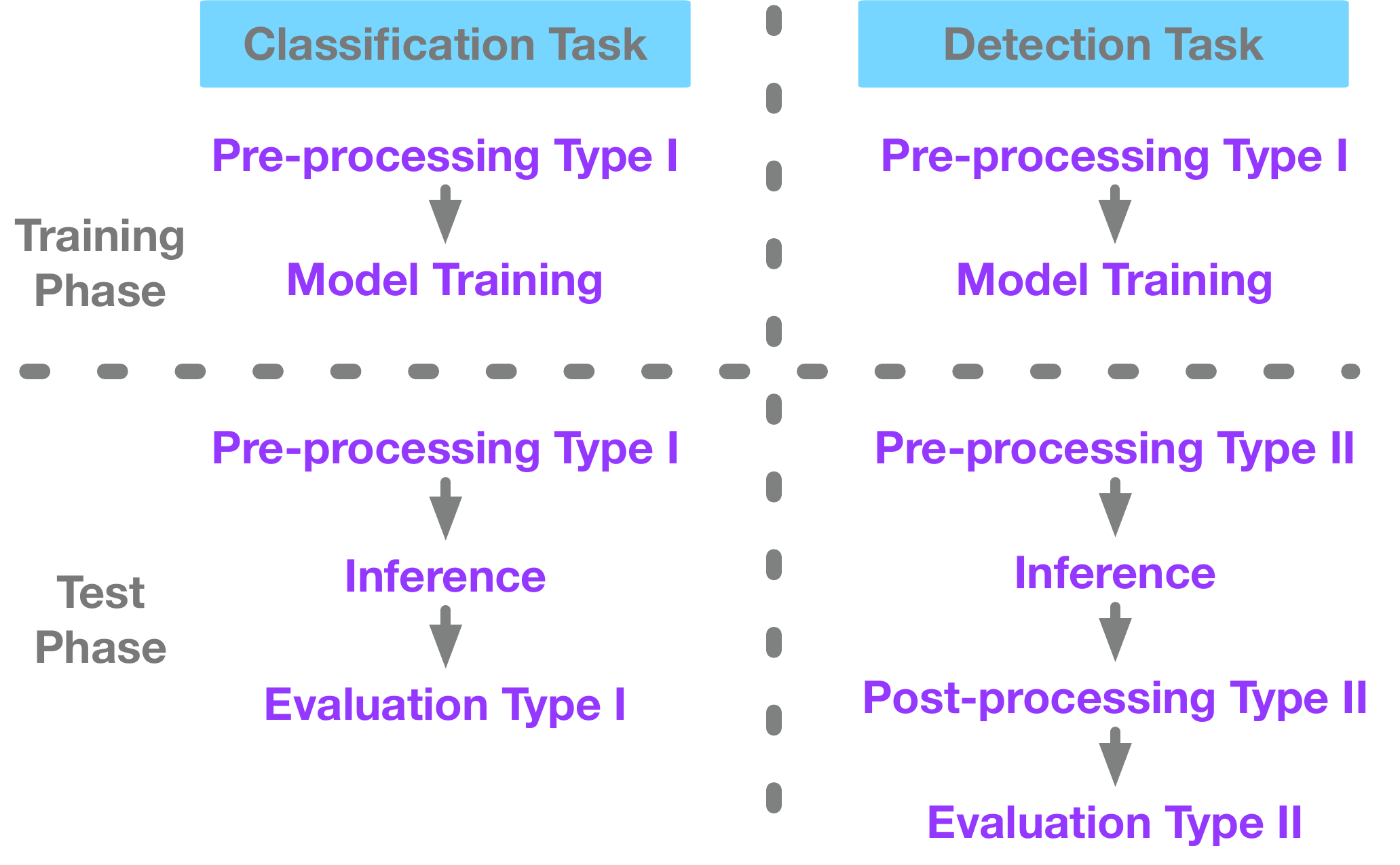}
    \caption{
   Task-based (Classification and Detection) pipelines summary.
    }
    \label{fig:Classif_Detec_pipeline}
\end{figure}

\subsubsection{Dataset 1}
We used Dataset 1 to train models for two tasks: seizure classification and seizure detection. Table 1 lists the processing stages, including data pre-processing, post-processing, and model evaluation.

\begin{figure}
    \centering
    \includegraphics[width=1\linewidth]{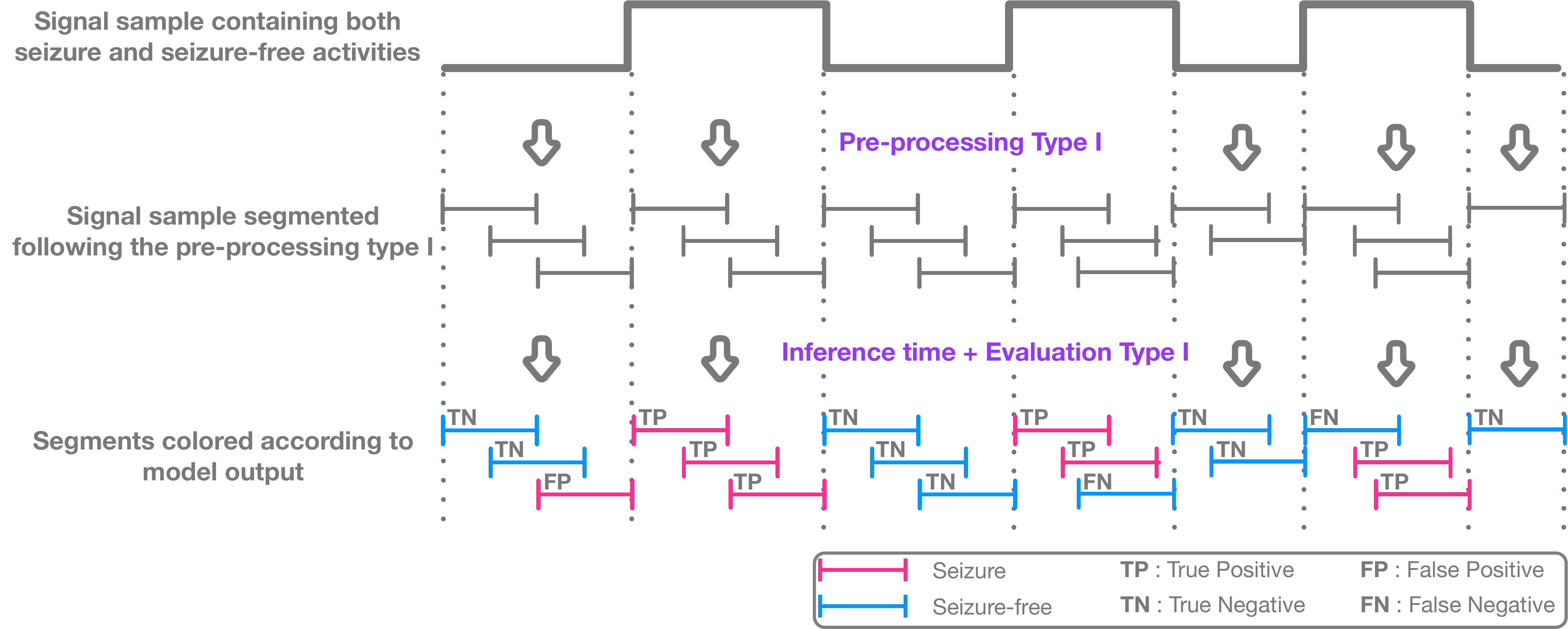}
    \caption{
    Illustration of the classification task pipeline at inference time. Segments built out of pre-processing I do not overlap across two activities. Segments colored in blue or red reflect an example of classification by the trained model. Blue color corresponds to classification into seizure-free activity. The red color indicates classification into seizure activity. \textbf{TP:} Segment labeled as seizure and detected as seizure by the model.\textbf{TN:} Segment labeled as seizure-free and detected as seizure-free. \textbf{FP:} Segment labeled as seizure-free and detected as seizure by the model. \textbf{FN:} Segment labeled as seizure and detected as seizure-free.
    }
    \label{fig:classification_pip}
\end{figure}

\paragraph{Pre-processing with prior identification of seizure/seizure-free activity (Pre-processing I)} We pre-process each EEG signal individually. Firstly, the signal is resampled from 512 Hz to 100 Hz. The resulting downsampled signal is filtered using a bandpass finite impulse response (FIR) filter between 1 and 20 Hz. We selected downsampling and filtering parameters based on their impact on the trained models’ accuracy. Then, we annotated continuous ranges of epileptic and seizure-free activity according to the onset and offset intervals of seizures labeled by experts. We then perform a Z-score normalization of the signal amplitudes. The mean and the variance for the normalization are calculated on the selection of all the amplitudes extracted from the seizure-free activity of the signal. Finally, we segment each range of seizure or seizure-free activities separately into 2- or 4-second blocks with variable overlap (or shift) size. The overlap is the common signal part between two consecutive blocks and the shift refers to the signal between the beginnings of two consecutive blocks. This pre-processing leads to 2- or 4-second blocks that do not contain mixed activities (seizure and seizure-free, please see Fig.\ref{fig:classification_pip}).

\begin{figure}
    \centering
    \includegraphics[width=1\linewidth]{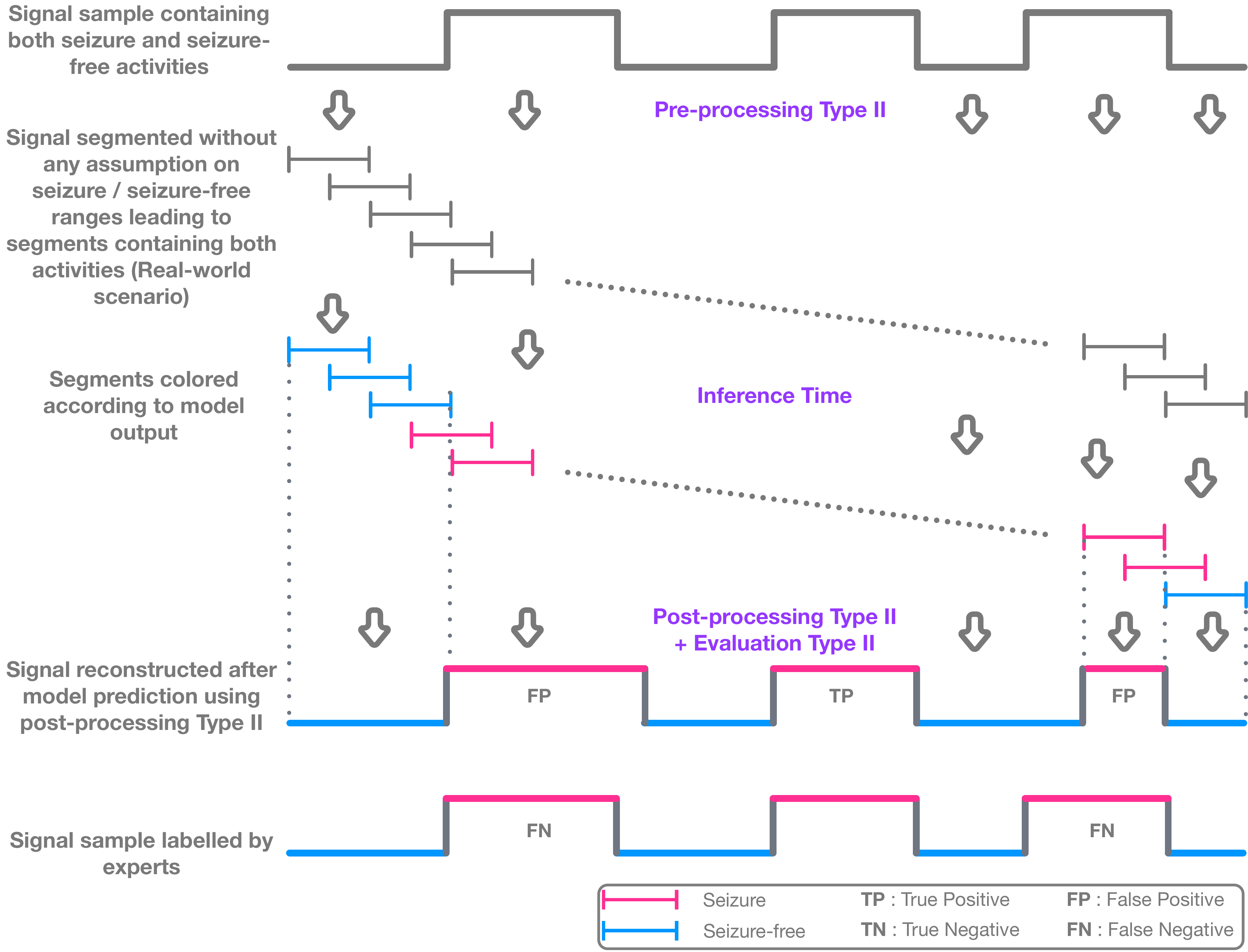}
    \caption{
       Illustration of the detection task pipeline at inference time. Segments built out of pre-processing II do overlap across activities exhibiting a real-world scenario. Segments predicted through the model application are re-assembled into a continuous signal using post-processing D. Finally, following evaluation II strategy, events formed in the reconstituted signal are compared with events labeled by the expert. \textbf{TP:} A seizure start \& end labeled by the expert matches a seizure start \& end detected by the model. \textbf{FP:} A seizure start \& end detected by the model did not find a match with any seizure start \& end labeled by the expert. \textbf{FN:} A seizure start \& end labeled by the expert did not find a match with any seizure start \& end detected by the model.
    }
    \label{fig:detection_pip}
\end{figure}

\paragraph{Pre-processing without prior identification of seizure/seizure-free activity (Pre-processing II)} Here, we also pre-process each EEG signal individually. The same resampling from 512 Hz to 100 Hz and bandpass filter from 1 to 20 Hz is applied. We then perform the Z-score normalization of the signal amplitudes as before. The mean and variance for the normalization are, however, computed using only the first 5 minutes of the signal. Finally, using a sliding window starting at time zero, the signal is segmented into overlapped 2- or 4-second blocks. It is worth noting that during this pre-processing procedure, the segmentation into 2- or 4-second blocks is applied without any prior distinction between seizure and seizure-free activities to mimic a real-world scenario. Such pre-processing will lead to 2- or 4-second blocks containing mixed activities (seizure and seizure-free, please see Fig.\ref{fig:detection_pip}).

\paragraph{Post-processing for signal reconstitution (Post-processing II)}
Following the application of a model at inference time to segments pre-processed without prior identification of seizure/seizure-free activity (Pre-processing II), we obtain a list of segments with labels predicted as seizures or non-seizures. Then, an ad-hoc post-processing algorithm is used to combine these segments and reconstruct the original signal. Segments with overlapping ranges are merged into a single range, and the predominant label is assigned to it. The procedure is iteratively applied until it produces a new set of labels on the continuous reconstructed EEG signal. We then compare this reconstructed signal to the original one following the evaluation II strategy.

\paragraph{Evaluation strategy for seizure classification (Evaluation I)}
After the model application at inference time to segments pre-processed with prior identification of seizure/seizure-free activity (Pre-processing I), we obtain a list of segments classified with seizure or seizure-free activity labels, one label per segment. Following a simple binary classification strategy, each predicted label is characterized as a correct or incorrect prediction, which allows the computation of true positives (TP: segment predicted as “seizure” while the true label is also “seizure”), true negatives (TN: segment predicted as “non-seizure” while the true label is also “non-seizure”), false positives (FP: segment predicted as “seizure” while the true label is “non-seizure”), and false negatives (FN: segment predicted as “non-seizure” while the true label is “seizure”).

\paragraph{Evaluation strategy for seizure detection (Evaluation II)}
This evaluation strategy aims to build a reliable metric for the seizure detection task. After signal reconstruction following the post-processing II method, we used an event-based metric previously introduced by \citep{mesaros2016metrics} to evaluate models. Adapted to the seizure detection task, this evaluation method compares each event (seizure) detected by the model to events labeled by the expert. \textbf{TP (True Positive) :} an event detected by the model overlaps an event labeled by the expert within a tolerance range of 1 second. The start of the detected event must be in a range of ± 1 second of the beginning of the labeled event and likewise for the end of the two events.
\textbf{FN (False Negative) :} an event labeled by the expert has not found a corresponding event detected by the model within the (+-) 1–second tolerance.
\textbf{FP (False Positive) :} an event detected by the model has not found a corresponding event labeled by the expert within the (+-) 1–second tolerance.

\subsubsection{Dataset 2}
Dataset 2 (Bonn public dataset) consists of signals already separated into seizure or non-seizure activities. Therefore, we could only perform a classification task on this dataset, as the detection task requires processing continuous signals containing both seizure and non-seizure activities.

To facilitate a comparison with Dataset 1, we also downsampled each signal of 23.6 seconds, from 173.61 Hz to 100 Hz. Then, a Z-score normalization is applied, using the mean and the variance calculated from sets A, B, C, or D, or the combination of some of them (refer to the subset description below). Initially, the Bonn dataset was unbalanced, with 100 segments (Set E) related to the seizure activity and 400 segments (Sets A, B, C, and D) labeled as seizure-free activity. To evaluate the robustness of our models, we introduced three balance ratios of the two classes by selecting different set combinations as follows:

\begin{itemize}
    \item \textbf{Subset 1:} a dataset constructed of sets A, B, C, D, and E. It is unbalanced, with 80\% of seizure-free activity and 20\% of seizure activity.
    
    \item \textbf{Subset 2:} a dataset constructed of sets A, B, and E. It is unbalanced, with 66.6\% of seizure-free activity and 33.3\% of seizure activity.
    
    \item \textbf{Subset 3:} a dataset formed of sets C and E. The balance is 50\% of seizure-free activity and 50\% of seizure activity.
    In all the assembled subsets, we used the segmentation into blocks of 4 seconds with 50\% overlap. We finally used the evaluation for seizure classification (Evaluation I) strategy to compute TP, TN, FP, and FN metrics.
\end{itemize}

\subsubsection{Training, Validation, and Test sets}
We applied a rigorous approach to data splitting to ensure data integrity and prevent data leakage in Dataset 1. Specifically, if a signal from animal A was allocated to the training set, we guaranteed that no other signals from the same animal A would be included in the validation or test sets.
Among the 136 animals, 100 were randomly selected to constitute the training set, 19 others for the validation set and 17 left for the test set. EEG signals recorded from the selected animals were assigned to each of the three groups.

\begin{itemize}
    \item \textbf{Training set:} EEGs recorded from the selected 100 animals correspond to 184 hours of epileptic seizure activity and 870 hours of seizure-free activity. To balance the training set, we selected 200 hours of seizure-free activity among the 870 hours.
    
    \item \textbf{Validation set:} EEGs recorded from the selected 19 animals correspond to 44 hours of epileptic seizure activity and 210 hours of seizure-free activity. To facilitate a more straightforward evaluation of the model during the training process, we selected 50 hours of seizure-free activity among the 210 hours to balance the validation set.
    
    \item \textbf{Test set:} EEGs recorded from the selected 17 animals correspond to 22 hours of seizure activity and 110 hours of seizure-free activity. We kept the test set unbalanced to evaluate the models’ generalization capabilities in a real-world scenario.
\end{itemize}

In this study, Dataset 2 is used entirely as a test set to evaluate the generalization capabilities of our best models.

\subsection{Network architectures}
\subsubsection{CNN-based architectures}
Convolutional neural networks excel in identifying local patterns in images or time series data \citep{roy2018deep}. The principal element constituting this network is a convolutional layer followed by a nonlinear activation function and, very often, resolution reduction operations like maximum/average pooling layers.

\paragraph{Classical CNN architectures} We constructed these network architectures by combining convolutional layers, batch normalization layers, Rectified Linear Unit (ReLU), max-pooling layers, SoftMax activation functions, and dense layers. We rigorously applied the following order to all the constructed CNN-based architectures. They all start with a convolutional layer followed by a batch normalization layer, a ReLU activation function and a max pooling layer. These blocks of four grouped layers are linked in a consecutive manner multiple times (3, 5, 6, 12, or 16). These blocks are followed by two dense layers separated by a ReLU function, and finally, the learned embeddings are followed by a SoftMax activation function for the final classification into epileptic/seizure-free activity. 

\paragraph{Customized U-Time architectures} U-Time \citep{perslev2019utime} is an architecture formed by a downstream network similar to a CNN-based network with blocks of convolutional, pooling, batch normalization layers, and activation functions. It is followed by an upstream network, also based on CNN-based networks but with pooling layers replaced by upsampling layers that increase the data dimensionality previously decreased by pooling layers on the downstream part. U-Time is a modified version of U-Net \citep{ronneberger2015unet}, a network where the term “U” refers to the shape of the network architecture. Such networks are called segmentation-based networks. The original U-Network allows us to perform image segmentation tasks. U-Time has been specifically adapted for time series data, like EEG signals. To adjust the U-Time network to our seizure detection task, we removed the classifier segment.

\subsubsection{CNN+RNN-based architecture}
Recurrent neural network (RNN) is a type of neural network characterized by a bi-directional flow. The output of some nodes combined with the following inputs of the same nodes guarantees a dependence between inputs and outputs as opposed to feed-forward networks like CNNs. Another distinguishable characteristic is that they share parameters across the layers in the network and have the same weight parameters within each layer, whereas feed-forward network types have different weights across each node. Typically, RNN-based networks are tailored to extract long-term relationships on time series data \citep{roy2018deep}. 
Classical RNN layers are prone to vanishing gradients. This limitation led to the development of long short-term memory (LSTM) layers \citep{sak2014lstm} and gated-recurrent unit (GRU) layers \citep{cho2014rnn}, inspired by classical RNN layers and less exposed to the gradient vanishing problem. We have combined CNN layers with either LSTM or GRU layers for our classification and seizure detection tasks.

\begin{figure}
    \centering
    \includegraphics[width=1\linewidth]{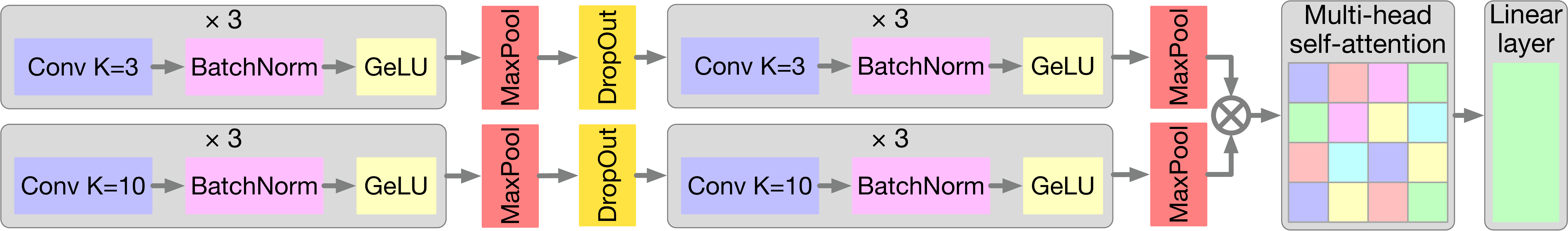}
    \caption{
    Schematic representation of the dual path CNN/transformer architecture. $K$ denotes the kernel size.
    }
    \label{fig:cnn_transformer_arch}
\end{figure}

\subsubsection{CNN+Transformer architecture (Fig.\ref{fig:cnn_transformer_arch})}
The transformer architecture described in \citet{vaswani2017attention} comprises an encoder and a decoder part. In our study, we only used the encoder part, as our tasks do not evolve into data generation. The encoder architecture contains a multi-head attention network, performing the computation of attention scores on pairs of EEG sequences and a position-wise fully connected feed-forward network. Each of the aforementioned networks is followed by a residual connection and a layer normalization.
Our CNN+Transformer architecture comprises a two-head attention network with embeddings from the raw EEG combined with positional encodings as inputs. The input embeddings are learned through two CNN-based architectures with six convolution blocks. The first CNN architecture contains convolution kernels of size 3, and the second is made of kernels of size 10 to capture frequency information on different scales \citep{eldele2021attention}. The positional encodings are built with the RoPE rotary position embedding introduced by \citet{su2023roformer}.

\subsection{Computational details}
We trained all the models for 100 epochs using the PyTorch framework and the Adam optimizer to optimize the learnable parameters. The hyperparameters optimization of the models (learning rate, number of epochs) was carried out using the validation set, and we specifically used the binary cross-entropy (BCE) loss function to tune the training performance. The computing resource used to train the models was an NVIDIA Tesla V100 GPU cluster with a single node containing 32 Go RAM.

\subsection{Performance metrics}

We considered the “seizure” state as the positive class and the “seizure-free” state as the negative class. Our main metrics are accuracy, sensitivity, precision, and F1-score.

\vspace{1em}

Accuracy indicates the ratio of correct predictions made by the model to the total number of predictions:
\begin{equation}
\text{Accuracy} = \frac{\text{Number of Correct Predictions}}{\text{Total Number of Predictions}}
% \tag{Eq. 1}
\end{equation}

\vspace{1em}

Sensitivity, also called Recall, is defined as the proportion of correctly categorized positive segments to all positive segments:
\begin{equation}
\text{Recall} = \frac{\text{True Positives}}{\text{True Positives} + \text{False Negatives}}
% \tag{Eq. 2}
\end{equation}

\vspace{1em}

Precision, the ratio of correctly predicted “seizure” segments to all segments predicted as “seizure”, is defined as:
\begin{equation}
\text{Precision} = \frac{\text{True Positives}}{\text{True Positives} + \text{False Positives}}
% \tag{Eq. 3}
\end{equation}

\vspace{1em}

Finally, F1-score is a metric that combines precision and recall to evaluate the model’s performance. It considers both false positives and false negatives:
\begin{equation}
\text{F1-score} = \frac{2 \times \text{Precision} \times \text{Recall}}{\text{Precision} + \text{Recall}}
% \tag{Eq. 4}
\end{equation}

%%%%%%%%%%%%%%%%%%%%%%%%%%%%%%%%%%%%%%%%%%%%%%%%%%%%%%%%%%%%%%%%%%%%%%%%
% \section{Results}
% \subsection{Dataset 1 models’ performance on seizure classification task}
% We evaluated the models trained on dataset 1 using the test set representing 10\% of the entire dataset. 
% Table 3 lists the success rates of the CNN-based architectures of different depths, from 3 to 16 convolutional layers. The model rendering the best performance is the one with 6 convolutional layers, which also appears to be the one containing the smallest number of parameters.

% \subsection{Dataset 1 models’ performance on seizure classification vs seizure detection tasks}

% \subsection{Models trained on dataset 1 and tested on dataset 2}

\section{Results}
%\label{sec:results}
This section reports results on two tasks and two datasets. On Dataset~1 (mice EEG), we compare CNNs, customized U-Time, CNN+RNN hybrids, and CNN+Transformer architectures for the segment-level seizure classification task, then contrast classification with event-level detection task, across different window/shift settings. Finally, we evaluate the generalization capability of our best architectures by training on Dataset~1 and testing on human EEG (Dataset~2). In each table, the best result according to the F1-score is highlighted in bold.

\subsection{Dataset~1: Architectural comparison for the seizure classification task}
%\label{subsec:cls_dataset1}

Across CNNs with 3--16 convolutional layers, the 6-layer CNN offers the best accuracy--capacity trade-off: it attains an F1-score of \textbf{0.818} with only \textbf{23k} trainable parameters (recall = 0.926), while going deeper or shallower does not yield consistent gains despite larger models. Customized U-Time variants underperform the CNN baselines (F1 = 0.656--0.726). Adding recurrent layers (biLSTM/GRU) to CNN-6 substantially increases capacity (2.6M parameters) without improving F1 (0.808--0.814). By contrast, coupling the 6-layer CNN with a Transformer encoder markedly improves performance to \textbf{F1 = 0.868} with \textbf{158k} parameters, indicating that attention captures temporal dependencies more efficiently than recurrent layers for this task (Table \ref{tab:cls_all}).

\begin{table}[ht]
\centering
\caption{Seizure classification task results on Dataset~1. All runs use 4~s windows with 2~s shift.}
\label{tab:cls_all}
\begin{tabular}{lrrr}
\hline
\textbf{Architecture} & \textbf{Params} & \textbf{Recall} & \textbf{F1-score} \\
\hline
CNN --- 3 layers             & 2.5 M  & 0.909 & 0.818 \\
CNN --- 5 layers             & 159 K  & 0.919 & 0.818 \\
CNN --- 6 layers             & 23 K & 0.926 & 0.818 \\
CNN --- 12 layers            & 216 K  & 0.918 & 0.816 \\
CNN --- 16 layers            & 258 K  & 0.907 & 0.807 \\
Customized U-Time --- 17 conv & 236 K & 0.876 & 0.656 \\
Customized U-Time --- 22 conv & 943 K & 0.900 & 0.726 \\
CNN-6 + biLSTM               & 2.6 M  & 0.888 & 0.814 \\
CNN-6 + GRU                  & 2.6 M  & 0.895 & 0.808 \\
\textbf{CNN + Transformer}   & \textbf{158 K} & \textbf{0.898} & \textbf{0.868} \\
\hline
\end{tabular}
\end{table}

\subsection{Seizure classification vs Seizure detection tasks}
%\label{subsec:det_dataset1}

We next evaluated the same architectures under the \emph{event-based detection} protocol (the seizure detection task). Different segment sizes (2, 4 seconds-long) associated with different overlaps (or shifts) have been tested.

As expected, detection is substantially harder than classification: F1-score values drop across the board (Table~\ref{tab:det_vs_cls}). Finer strides consistently help---\textbf{0.5~s} shifts yield higher detection F1 than 1 or 2~s. The \textbf{CNN+Transformer} remains the top performer, peaking at \textbf{F1 = 0.565} with 4~s windows and 0.5~s shifts; with 2~s windows it reaches F1 = 0.529 (0.5~s shift) and 0.481 (1~s shift). In contrast, CNN-only and CNN+RNN models degrade sharply under detection task, reflecting difficulties with boundary localization once segments can contain mixed activities. Taken together, these findings underscore two points: (i) segment-level classification task can overestimate real-world detection capability, and (ii) denser overlaps improve event reconstruction and boundary precision in the detection task.

\begin{table*}[t]
\centering
\caption{Comparison of model performance on \emph{classification} vs. \emph{detection} tasks on Dataset~1. We report two types of window/shift settings for classification and four types for detection.}
\label{tab:det_vs_cls}
\begin{tabular}{lcc|cccc}
\hline
\multirow{2}{*}{\textbf{Architecture}} & \multicolumn{2}{c|}{\textbf{F1-score (Classification)}} & \multicolumn{4}{c}{\textbf{F1-score (Detection)}} \\
\cline{2-3}\cline{4-7}
 & 4~s / 2~s & 2~s / 1~s & 4~s / 2~s & 4~s / 0.5~s & 2~s / 1~s & 2~s / 0.5~s \\
\hline
CNN --- 6 layers         & 0.818 & 0.722 & 0.000 & 0.182 & 0.381 & 0.496 \\
CNN-6 + biLSTM           & 0.814 & 0.725 & 0.000 & 0.160 & 0.242 & 0.377 \\
Customized U-Time        & 0.726 & 0.694 & 0.000 & 0.000 & 0.000 & 0.000 \\
\textbf{CNN + Transformer}      & \textbf{0.868} & \textbf{0.818} & \textbf{0.363} & \textbf{0.565} & \textbf{0.481} & \textbf{0.529} \\
\hline
\end{tabular}
\end{table*}

\subsection{Train on Dataset~1 (mice) and test on Dataset~2 (humans)}
%\label{subsec:cross_dataset}

One goal of preclinical studies is to enhance risk management and anticipate outcomes on future human experiments. Following this motivation, we trained our two best architectures (CNN-6 and CNN+Transformer) on mice EEG (Dataset~1) and evaluated them on human EEG (Dataset~2; Bonn subsets). Results are shown in Table~\ref{tab:bonn_cross}. The \textbf{CNN+Transformer} demonstrates strong robustness and generalization: on unbalanced Subsets~1--2 it achieves F1 = 0.888 and 0.852, respectively, while \textbf{CNN-6} exhibits very high recall (0.952--0.956) but modest F1 (0.627--0.638), indicating many false positives. On the balanced Subset~3, both models improve; CNN+Transformer reaches \textbf{F1 = 0.935} (recall = 0.893) and CNN-6 reaches F1 = 0.908 (recall = 0.950). These results suggest that attention-based modeling translates seizure-related signatures learned in mice to humans more robustly than CNNs alone, especially under class imbalance.

\begin{table*}[t]
\centering
\caption{Performance of our two best architectures when trained on Dataset~1 (mice) and evaluated on Dataset~2 (humans; Bonn). Subset definitions: \emph{Subset~1}: Seiz = Set~E; Non-seiz = Sets~A,B,C,D. \emph{Subset~2}: Seiz = Set~E; Non-seiz = Sets~A,B. \emph{Subset~3}: Seiz = Set~E; Non-seiz = Set~C.}
\label{tab:bonn_cross}
\begin{tabular}{lcccccc}
\hline
\multirow{2}{*}{\textbf{Model}} & \multicolumn{2}{c}{\textbf{Subset 1}} & \multicolumn{2}{c}{\textbf{Subset 2}} & \multicolumn{2}{c}{\textbf{Subset 3}} \\
\cline{2-7}
 & Recall & F1-score & Recall & F1-score & Recall & F1-score \\
\hline
CNN --- 6 layers      & 0.956 & 0.627 & 0.952 & 0.638 & 0.950 & 0.908 \\
\textbf{CNN + Transformer}     & \textbf{0.904} & \textbf{0.888} & \textbf{0.896} & \textbf{0.852} & \textbf{0.893} & \textbf{0.935} \\
\hline
\end{tabular}
\end{table*}

%%%%%%%%%%%%%%%%%%%%%%%%%%%%%%%%%%%%%%%%%%%%%%%%%%%%%%%%%%%%%%%%%%%%%%%%
\section{Discussions and perspectives}
The application of machine learning algorithms, including deep learning neural networks, has witnessed significant progress in detecting seizure activity from electroencephalography (EEG) recordings over the past few decades. This advancement holds promise for enhancing clinical treatment outcomes and deepening our understanding of the underlying neurobiological mechanisms. Notably, since the early 90s, numerous studies have consistently demonstrated the ability of machine learning to identify seizures with high sensitivity, typically exceeding 95\% \citep{jando1993pattern,medvedev2024absence}. However, these findings often have elevated rates of false positive detections. As a result, they lack generalization across subjects, highlighting the need for continued refinement of these approaches.

\subsection{Challenges of dataset size and feature engineering}
The increasing use of deep learning techniques for analyzing raw EEG data has been limited by the small size of many EEG datasets, which poses challenges for training reliable models. Collecting large datasets can be both resource-intensive and time-consuming, often exceeding the capabilities of smaller research centers. Previously, the standard approaches involved manually creating features and applying machine learning or deep learning methods alongside traditional explainability techniques \citep{chen2023automated, ellis2023augmentation, ince2008selection, kwon2018emotion, ruffini2019deep}. These manually extracted features typically captured aspects of the data in the time domain and/or frequency domain. While effective, these methods are inherently restricted by the limited feature space available for learning. In contrast, deep learning methods, such as CNNs, can automatically learn features by creating robust embedding spaces, making them an appealing solution for raw EEG analysis \citep{oh2019deep, shoeibi2021automatic}. In the present study, we leverage the availability of a large dataset collected under consistent conditions (hardware and recording parameters) in an animal model of MTLE, allowing effective use of CNNs and transformers with raw electrophysiological signals.

\subsection{Preventing data leakage in EEG-based studies}
One of the key strengths of this study lies in its consideration of the potential for data leakage during dataset training. A common pitfall in EEG-based studies is the random assignment of segments to training and test sets, which results in data samples from individual subjects being parts of both sets. Such assignments can lead to data leakage, where EEG segments from a single subject appear in both the training and test sets, thus artificially inflating model performance. A recent study by \citet{brookshire2024leakage} highlighted the importance of addressing this issue by comparing the performance of deep neural network (DNN) classifiers using segment-based holdout (where segments from one subject can appear in both sets) versus subject-based holdout (where all segments from one subject are exclusive to either the training or test set). The authors demonstrated that segment-based holdout can lead to a significant overestimation of the model's performance on previously unseen subjects. Alarmingly, they found that most translational DNN-EEG studies employ segment-based holdout, which may result in a dramatic overestimation of the model's performance on new subjects \citep{rasheed2021review,shoeibi2021automatic}. To ensure an accurate assessment of our model, we designed a rigorous approach by exclusively including each subject's data to only the training or test sets, but never both.

\subsection{Limitations of conventional pre-processing and evaluation strategies}
All existing methods for seizure detection involve a pre-processing step that separates seizure activities from non-seizure activities. This process typically segments the data into small blocks to create training and test sets. As noted, this pre-processing technique lead to an ideal scenario where each small block contains either only seizure activities or only seizure-free activities. In this study, we demonstrate that such a pre-processing pipeline tends to overestimate the performance of seizure detection. To better reflect the real-world challenges associated with automatic seizure detection, it is essential to develop pipelines without any prior differentiation between seizure and seizure-free activities (see the Pre-processing II method). Furthermore, models should primarily be evaluated based on their ability to accurately detect the onset and offset of seizures in continuous EEG signals (refer to the Post-processing II and Evaluation II methods). The Post-processing II algorithm reconstituted continuous EEG signals from overlapping segments, maintaining a temporal resolution of 500 ms. This improvement increased the model's precision and facilitated comparisons between the predicted and labeled onset/offset of seizures. The analysis of two distinct evaluation strategies (Evaluation I and Evaluation II) highlights the fundamental differences between the classification and detection tasks of seizures in EEG signals.

\subsection{Robustness and trans-species adaptability of our approach, Futur directions}
One of the most exciting aspects of this study is the robustness of our approach, which yields comparable performance metrics to some commercial systems \citep{koren2021seizure}, even when faced with modifications to the recording setup, changes in recording conditions, or differences in environment (experimental/clinical) or species (mouse/human) \citep{besne2022interactive}. This versatility underscores the potential of our approach to transcend traditional boundaries and facilitate seamless translation between preclinical research and clinical applications. Our results demonstrate the capabilities of the proposed approach in both clinical and research environments, offering a valuable tool to aid experts in alleviating the burden of annotating extensive hours-long EEG recordings. Furthermore, the trans-species adaptability of our approach may facilitate a deeper understanding of the differences and similarities between human diseases and animal models. Notably, to the best of our knowledge, this study is one of the few to validate a high-performance detection algorithm for HPDs on comprehensive EEG datasets from both animal (mice MTLE dataset) and human (Bonn dataset) subjects. The detection performance of our proposed method suggests that this approach can be reliably applied in preclinical research and clinical settings, paving the way for future studies to explore its potential in real-world applications. Future work should focus on validating this framework using extended EEG data with diverse seizure types to evaluate its specificity and expand its applications. Additionally, it would be beneficial to validate our pipeline on other animal models to enable its broader use in preclinical research. Although our methodology demonstrates promising results in terms of generalization capabilities, it would be valuable to apply explainability methods to gain insights into the features learned by the model. Our approach, which utilizes deep neural networks trained directly on raw EEG data, complicates this objective. This presents a significant disadvantage compared to pipelines that rely on models trained with extracted features, where relatively straightforward studies can effectively highlight the contributions of various features to seizure detection \citep{sturm2016interpretable}.

%%%%%%%%%%%%%%%%%%%%%%%%%%%%%%%%%%%%%%%%%%%%%%%%%%%%%%%%%%%%%%%%%%%%%%%%
\section{Conclusion}

This work introduces a novel seizure detection pipeline exhibiting pre-processing and post-processing techniques relevant to real-world scenarios. The experiments conducted revealed that neglecting these real cases can lead to an overestimation of the models’ performance. We implemented several architectures for the seizure detection task, including CNN, RNN, segmentation, and transformer-based models. Our experiments demonstrated that the best-performing architecture, which combines a CNN with a transformer encoder, exhibits strong generalization abilities. This model was trained on raw EEG signals from animals and tested on raw EEG data from humans, showing strong generalization capabilities.

\paragraph{CRediT  authorship contribution statement}
Davy Darankoum: Writing – review \& editing, Writing – original draft, Visualization, Validation, Software, Methodology, Investigation, Formal analysis, Data curation, Conceptualization. Manon Villalba: Investigation. Clélia Allioux: Investigation. Baptiste Caraballo: Investigation. Carine Dumont: Investigation. Eloïse Gronlier: Investigation. Corinne Roucard: Writing - Review \& Editing, Project administration. Yann Roche: Writing - Review \& Editing, Funding acquisition. Chloé Habermacher: Writing - Review \& Editing, Validation. Julien Volle: Writing - Review \& Editing, Supervision, Conceptualization. Sergei Grudinin: Writing - Review \& Editing, Supervision, Conceptualization.

\paragraph{Funding statement}
This work was funded by SynapCell SAS through the Cortex project which has been awarded at the 9th edition of the i-Nov competition organized for French companies. The research has been conducted using both SynapCell and Laboratoire Jean Kuntzmann (CNRS/UGA) resources.

\paragraph{Declaration of Competing Interest}
Davy Darankoum, Manon Villalba, Clélia Allioux, Baptiste Caraballo, Carine Dumont, Eloïse Gronlier, Corinne Roucard, Yann Roche, Chloé Habermacher, and Julien Volle are employees of SynapCell SAS. Sergei Grudinin have no conflicts of interest to declare.

\paragraph{Data availability statement}
Dataset 1 can be made available to independent researchers after receipt of a valid research proposal, data analysis plan, and summary of researcher qualifications. Requests may be submitted to SynapCell at \url{https://synapcell.com/contact-us/}. Provision of data is contingent on business feasibility and execution of a data use agreement.

The Bonn EEG time series (Dataset 2) is available at \url{https://repositori.upf.edu/handle/10230/42894}.

% Dataset 2 is available at 
% \href{https://www.upf.edu/web/ntsa/downloads/-/asset_publisher/xvT6E4pczrBw/content/2001-indications-of-nonlinear-deterministic-and-finite-dimensional-structures-in-time-series-of-brain-electrical-activity-dependence-on-recording-regi?_com_liferay_asset_publisher_web_portlet_AssetPublisherPortlet_INSTANCE_xvT6E4pczrBw_assetEntryId=229569389&_com_liferay_asset_publisher_web_portlet_AssetPublisherPortlet_INSTANCE_xvT6E4pczrBw_redirect=https%3A%2F%2Fwww.upf.edu%3A443%2Fweb%2Fntsa%2Fdownloads%3Fp_p_id%3Dcom_liferay_asset_publisher_web_portlet_AssetPublisherPortlet_INSTANCE_xvT6E4pczrBw%26p_p_lifecycle%3D0%26p_p_state%3Dnormal%26p_p_mode%3Dview%26_com_liferay_asset_publisher_web_portlet_AssetPublisherPortlet_INSTANCE_xvT6E4pczrBw_cur%3D0%26p_r_p_resetCur%3Dfalse%26_com_liferay_asset_publisher_web_portlet_AssetPublisherPortlet_INSTANCE_xvT6E4pczrBw_assetEntryId%3D229569389}{"link"}.

\clearpage
%%%%%%%%%%%%%%%%%%%%%%%%%%%%%%%%%%%%%%%%%%%%%%%%%%%%%%%%%%%%%%%%%%%%%%%%

%%% Use this command to include your bibliography file.

%\bibliography{Seiz_Detec}
%\bibliographystyle{unsrtnat}

\end{document}